\documentclass[conference]{IEEEtran}
\IEEEoverridecommandlockouts
\usepackage{cite}
\usepackage{amsmath,amssymb,amsfonts}
\usepackage{algorithmic}
\usepackage{graphicx}
\usepackage{textcomp}
\usepackage{xcolor}
\usepackage{colortbl}
\usepackage{url}
\usepackage{subfig}
\usepackage{booktabs}
\usepackage{multirow}

\renewcommand{\footnoterule}{%
  \kern -3pt
  \hrule width 0.75\columnwidth height 0.4pt
  \kern 2.6pt
}
\usepackage[skip=5pt, position=bottom]{caption}

\begin{document}

\title{Beyond Channel Data: Configuration-Enriched Radio Foundation Models}
\author{\IEEEauthorblockN{Shrihari Vasudevan\IEEEauthorrefmark{1}, H. G. Ranjani\IEEEauthorrefmark{1}, Raaghul Ranganathan\IEEEauthorrefmark{1}, Satwik Sahu\IEEEauthorrefmark{2}, Thirumaran Ekambaram\IEEEauthorrefmark{1}}
\IEEEauthorblockA{\IEEEauthorrefmark{1}
\textit{Ericsson R\&D, India}, \IEEEauthorrefmark{2} {BITS Pilani, India}\\
\IEEEauthorrefmark{1}\{shrihari.vasudevan, ranjani.h.g, r.raaghul, thirumaran.ekambaram\}@ericsson.com, \IEEEauthorrefmark{2} stwk17@gmail.com}}

\maketitle

\begin{abstract}

This paper proposes incorporating channel configuration ($\text{CFG}$) information into Radio
Foundation Model (RFM) pre-training and demonstrates that the resulting $\text{CFG}$-enriched embeddings
serve as effective inductive priors for downstream tasks under label scarcity. We compare the proposed embeddings with raw channel tokens and multiple baseline models, under three different data regimes, for multiple downstream tasks --- Line-of-Sight/Non-Line-of-Sight (LoS/NLoS) classification, timing advance (TA) estimation, and beam selection. Evaluation is done in both in-distribution and out-of-distribution configuration settings. Experimental outcomes  show that the proposed embeddings consistently outperform and/or are competitive to the baselines, across experiments in both seen and unseen configuration scenarios. 

\end{abstract}


\section{Introduction}
In any receiver, the channel frequency response (CFR), $H$ is estimated from pilot signals and subsequently used to decode the input bitstream. In addition, it may be leveraged for a range of downstream tasks including positioning, beam management, and channel prediction. Radio foundation models (RFM) bring recent advances of machine learning to wireless data; RFMs have been developed using both the CFR, i.e., $H$ \cite{LWM, CSICLIP, WiFo, WiFo2, ContraWiMAE} and the received signal, $Y$ \cite{IQFM, LWMSpectro}. Typically, RFMs are trained on unlabeled $H$ (and/or $Y$) via self-supervised objectives, producing their respective embeddings. Existing works in literature demonstrate improved generalization across multiple downstream tasks, with minimal task-specific labeled $H$ (and/or $Y$) data which is reflective of real world scenario. This paper is about the input data for RFMs; specifically, it proposes incorporating the channel configuration (referred to as $\text{CFG}$) into the token sequence used to develop the RFM. A CFR-based RFM is used as an example while the hypothesis of this paper is more broadly applicable. 

\subsection{Related work}
\label{sec:related_work}

Recent AI-for-RAN works adapt advances in NLP and Computer Vision to represent wireless channels. Transformers dominate as the RFM backbone; some are encoder-centric designs with light masking and a Masked Channel Modeling (MCM) objective \cite{LWM}, and others use decoder-centric designs with aggressive masking and a Masked Auto-Encoder (MAE) objectives \cite{WiFo, WiFo2, ContraWiMAE, PilotWiMAE}.  A dual ResNet50 encoder architecture coupled with a contrastive InfoNCE (Information Noise-Contrastive Estimation) loss was proposed in \cite{CSICLIP}, to align time and frequency domain representations. Some works combine objectives - as examples, \cite{ContraWiMAE} merges MCM with contrastive learning via noisy masked reconstruction and \cite{LWLM} uses a multi-objective loss combining MCM, contrastive, and domain-invariance terms. The work \cite{FilterAndAttend} extends \cite{LWM} with noise and interference suppression. A Mixture of Experts (MoE) architecture within the Transformer layers was proposed in \cite{WiFo2, WiFoE}, the latter learning pre-coding strategies rather than direct channel representations. This paper builds on \cite{LWM} and quantifies improvements obtained by incorporating $\text{CFG}$ alongside $H$ - a hypothesis broadly applicable to all cited works.

Most works, including this paper, use the DeepMIMO dataset\footnote{\url{https://deepmimo.net/}}. Majority of the relevant literature are based on DeepMIMO dataset and consider a standard 32 antennas with 32 subcarriers configuration for generating $H$. Other tools for $H$ include NVIDIA Sionna \cite{sionna} and Quadriga \cite{quadriga}. A recent benchmark \cite{CFMBench} unifies multiple channel generation mechanism and provides complementary multi-domain comparison of channel RFMs. Works such as \cite{WiFo, WiFo2, WiFoE} handle configuration diversity via variable-length tokenization and MoE routing. However, except \cite{LWLM}, which incorporates Base Station (BS) position in the decoder for a positioning objective, none of the existing works explicitly encodes channel generation $\text{CFG}$ into the RFM. 

This paper proposes and demonstrates that augmenting CFR with configuration information (i.e., $H$ with $\text{CFG}$) helps in generalization abilities, specifically in out-of-distribution (OOD) scenarios. Experiments in this paper use CFR data from DeepMIMO along with the $\text{CFG}$ that are used for generating the same. The embeddings of the configuration information are assessed by adopting the probing methodology from literature \cite{ProbingSentEmb, BERTPipeline, ViTExploration} i.e., evaluating frozen embeddings on multiple downstream tasks across varying training-data sizes, to characterize both overall efficacy and data-efficiency regimes.

\subsection{Research Questions} 
As mentioned earlier, none of the existing work leverage $\text{CFG}$ information during training of RFM, referred to as the pretraining phase. Consequently, this work aims to answer the following research questions:
\begin{itemize}
    \item assessing the impact of configuration information enrichment on baseline/existing embeddings
    \item understanding the efficacy of the newly available $\text{CFG}$-enriched embeddings -  using different training sizes for downstream tasks and evaluating on both seen and OOD scenarios.
\end{itemize}


\subsection{Contributions}
The following are the contributions of this work:
\begin{itemize}
    \item demonstrates that the RFM embeddings vastly outperform raw data with or without raw CFG augmentation,
    \item demonstrates that encoding $\text{CFG}$ along with $H$ during RFM pretraining results in competitive to better downstream task performance; its positive impact is more often visible in OOD scenarios, suggesting that the $\text{CFG}$ information being encoded, provides an inductive prior, for more effective downstream generalization, and
    \item demonstrates the viability of $\text{CFG}$-embeddings, as a low-dimensional alternative to other lighter (e.g., CLS) or heavier $H$ embeddings.
\end{itemize}

The work is organized as follows: The details of the proposed approach to introduce configurations alongside channel are described in Section \ref{sec:proposed_approach}. The experimental setup along with results and observations are detailed in Section \ref{sec:experiments}. The work concludes with a summary of insights from these experiments. 

\section{Proposed Approach}
\label{sec:proposed_approach}

\subsection{Architecture}
The paper builds on the transformer architecture from \cite{LWM} to demonstrate the idea put forth; the hypothesis of this paper is however applicable for any RFM architecture. A multi-layer, multi-head Attention (MHA), Transformer Neural Network is used for the channel RFM encoder. The decoder is a simple fully-connected network. 

The complex CFR, $H$ is tokenized by extracting non-overlapping patches of size $ps$ from the flattened real and imaginary components. The input sequence $T_{\text{CLS}, H}$ consists of a $\text{CLS}$ token followed by $n/2$ real and $n/2$ imaginary tokens, where $n$ is the maximum sequence length and each token is a real-valued vector of length $ps$. The baseline RFM $f^b: \mathbb{R}^{(1+n) \times ps} \to \mathbb{R}^{d}$ maps this token sequence to $d$ dimensional embeddings:
\begin{equation}
    E^b = f^b(T_{\text{CLS}, H};\theta^b)
\end{equation}

where $\theta^b$ are the pretrained weights (the process of learning these weights is called pretraining). Specific embedding strategies ($\text{CLS}$, mean pool, max pool) are denoted $E^b_{\text{CLS}}$, $E^b_{H_\text{mean}}$, $E^b_{H_\text{max}}$ respectively, where $E^b_{\text{CLS}}$ is the output at the $\text{CLS}$ token position after joint processing of the full sequence ($T_{\text{CLS}, H}$) through the transformer, implicitly encoding global $H$ context via self-attention, and $E^b_{H_\text{mean}}$, $E^b_{H_\text{max}}$ are the mean and max pooled embeddings over the $n$ channel tokens.


To incorporate $\text{CFG}$ into the RFM, $\text{CFG}$ tokens are introduced between the $\text{CLS}$ and $H$ tokens, forming the input token sequence $T_{\text{CLS}, \text{CFG}, H}$. Each $\text{CFG}$ token is a normalized, $ps$-padded real-valued vector, with $\text{CFG} \in \mathbb{R}^m$. The proposed RFM $f^p: \mathbb{R}^{(1+m+n) \times ps} \to \mathbb{R}^{d}$ maps this augmented sequence to embeddings:
\begin{equation}
    E^p = f^p(T_{\text{CLS}, \text{CFG}, H};, \theta^p)
\end{equation}

where $\theta^p$ are the pretrained weights. The resulting embedding strategies are denoted $E^p_{\text{CLS}^\dagger}$, $E^p_{\text{CFG}_\text{mean}}$, $E^p_{\text{CFG}_\text{max}}$, $E^p_{H,\text{CFG}_\text{mean}}$, $E^p_{H,\text{CFG}_\text{max}}$, corresponding to the $\text{CLS}$ token output, mean/max pooling over embeddings of $\text{CFG}$ and $\text{CFG}$-enriched channel tokens, respectively. We also include $E^p_{H, \text{CLS},\text{CFG}_\text{mean}}$, $E^p_{H, \text{CLS},\text{CFG}_\text{max}}$ to evaluate pooling across the entire sequence. Details are listed in \ref{sec:experiments}.

\begin{figure}[htbp]
\centerline{\includegraphics[width=\columnwidth]{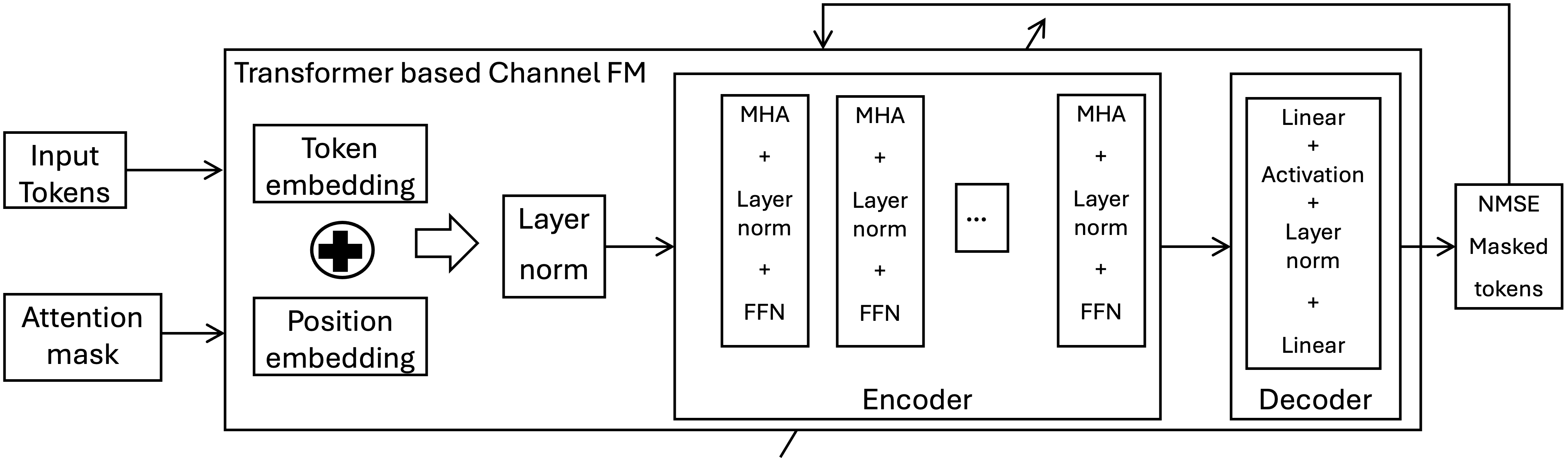}}
\caption{Baseline channel RFM. The Encoder comprises multiple Transformer layers with MHA, Layer normalization and FFN (Feed-forward Network). The Decoder consists of one linear layer. NMSE - Normalized Mean Squared Error. 
} 
\label{fig:model_arch}
\end{figure}

\begin{figure}[htbp]
\centerline{\includegraphics[width=\columnwidth]{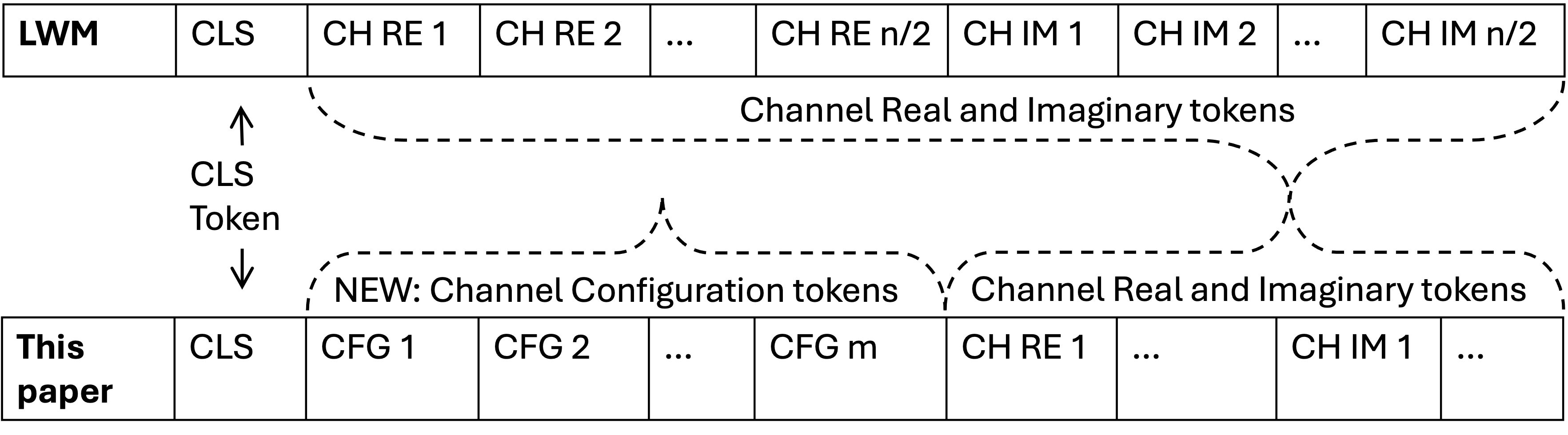}}
\caption{Architecture of proposed work. In comparison to baseline, this paper introduces $\text{CFG}$ tokens along with $H$ and $CLS$ tokens. 
}
\label{fig:diff_tokenization}
\end{figure}

 
 A simplified representation of the baseline CFR-based RFM architecture is shown in Figure \ref{fig:model_arch}. Tokenization difference between baseline and this paper is shown in Figure \ref{fig:diff_tokenization}. Specific $\text{CFG}$ data used in this paper are detailed in Section \ref{sec:experiments} and Table \ref{tab:pretraining_configs}. 

Post tokenization, the rest of the architecture is similar to that of the baseline. The details are listed in Table \ref{tab:hp}. While the baseline implementation starts with the codebase \cite{LWM}, it is modified to incorporate: (1) Gaussian Error Linear Unit (GELU) as activation function (2)  Weighted Normalized Mean Squared Error (NMSE) as a loss function; the weighting is to ensure balanced treatment of different scenarios, (3) early-stopping (4) learning-rate warm-up mechanism, and (5) adaptive gradient clipping\footnote{This paper adopts an adaptive gradient clipping approach to stabilize gradients and prevent pretraining collapses by using rolling median of maximum gradient norms across all layers.}. 

\subsection{Hypothesis}

The CFR, $H$ is governed by transmitter (Tx) and receiver (Rx) medium properties - antenna geometry, carrier frequency, bandwidth, and subcarrier spacing - collectively captured by $\text{CFG}$ tokens, which abstracts channel data across antenna elements and potentially neighboring Tx-Rx pairs. $\text{CFG}$ embeddings, initialized via a learnable linear projection, augment the random channel embeddings prior to pretraining.


The MCM objective trains the RFM to predict masked $H$ tokens from visible ones, capturing inter-token correlations. Introducing $\text{CFG}$ tokens extends this to learning $H-\text{CFG}$ correlations, such that $E_{H, \text{CFG}}$ embeddings encode both channel and configuration patterns with $\text{CFG}$ as an explicit prior. This paper hypothesizes that $E_{\text{CFG}}$ forms an effective, low-dimensional inductive prior for downstream generalization — particularly in (i) low labeled data regimes (ii) unseen configurations. To test this, (mean and max) pooled embeddings corresponding to $E^p_{\text{CFG}}$ and $E^p_{H,\text{CFG}}$ are compared against those of $E^b_H$ and $E^b_\text{CLS}$ - {\color{black} $E^p_{CLS}$, $E^b_{CLS}$}. In addition, we baseline the outcomes with the performance of an FCN model with $T_H$ as input sequence and $T_{H, \text{CFG}}$-augmented variants of the channel token sequence across tasks and training data sizes.

\section{Experiments And Results}
\label{sec:experiments}
\subsection{Pretraining configurations}
The two RFMs ($f^b$ and $f^p$) are pretrained using eight DeepMIMO scenarios\footnote{Pretraining scenarios - \texttt{city\_9\_sanfrancisco\_3p5\_lwm}, \texttt{city\_17\_seattle\_3p5}, \texttt{city\_7\_sandiego\_3p5\_lwm}, \texttt{city\_10\_florida\_villa\_7gp\_1758096510611}, \texttt{city\_10\_florida\_villa\_7gp\_1758095156175}, \texttt{city\_10\_florida\_villa\_7gp\_1758092194311}, \texttt{city\_10\_florida\_villa\_7gp\_1758026450854}, \texttt{city\_01\_villas\_florida\_1gp}.} i.e., $\text{CFG}$ tokens are not included in baseline model ($f^b$) and is included in proposed model, ($f^p$); refer Section \ref{sec:proposed_approach}. Data diversity is simulated for pretraining using diverse set of configurations; refer Table \ref{tab:pretraining_configs}. A pre-defined, empirically chosen set of architectural and pretraining hyperparameters are used, captured in Table \ref{tab:hp}.

\begin{table}[htbp]
    \caption{Pretraining scenario configurations resulting in $\sim$126K samples of $H$ from ~30 unique $\text{CFG}$ variants. }
    \begin{center}
    \begin{tabular}{|l|c|}
    \hline
    \textbf{Parameter}&\textbf{Pretraining} \\
    \hline
    Carrier frequency          & 3.5 GHz \\
    \hline
    {Subcarriers (OFDM)}        & 512 \\
    \hline
    {Selected subcarriers}$^{\mathrm{a}}$   & 48, 96, 144, 192, 288, 384, or 480 \\
    \hline
    {BS antenna shape}$^{\mathrm{a}}$ & [4,2], [4,4], [6,4], [8,4], or [8,8] \\
    \hline
    {UE antenna shape}$^{\mathrm{a}}$ & [2,1] or [2,2] \\
    \hline
    {BS antenna spacing} & 0.25$\lambda$ \\
    \hline
    {UE antenna spacing}$^{\mathrm{a}}$ & 0.25$\lambda$ or 0.5$\lambda$ \\
    \hline
    \multicolumn{2}{l}{$^{\mathrm{a}}$Randomized per {\color{black}BS} from the indicated options.} \\
    \end{tabular}
    \label{tab:pretraining_configs}
    \end{center}
\end{table}
      
\begin{table}[htbp]
  \caption{Architecture and pretraining Hyperparameters}
  \begin{center}
  \begin{tabular}{|l|c||l|c|}
  \hline
  \multicolumn{2}{|c||}{\textbf{Architecture}} & \multicolumn{2}{c|}{\textbf{Pretraining}} \\
  \hline
  Embedding dim, $d$ & 64 & Optimizer & AdamW \\
  \hline
  Layers & 12 & Learning rate & 1e-4 \\
  \hline
  Attention heads & 12 & Batch size & 16 \\
  \hline
  Activation & GELU & Weight decay & 1e-5 \\
  \hline
  Patch size, $ps$ & 16 & Warm-up steps & 50 \\
  \hline
  Dropout & 0.1 & Clip multiplier & 2.0 \\
  \hline
  Max input tokens, $n$ & 1729$^{\mathrm{b}}$ / 1735$^{\mathrm{p}}$ & Clip window & 5 \\
  \hline
  \multicolumn{4}{l}{$^{\mathrm{b}}$Baseline - CLS + 1728 channel tokens} \\
  \multicolumn{4}{l}{$^{\mathrm{p}}$Proposed - CLS + 6 Config + 1728 channel tokens} \\
  \end{tabular}
  \label{tab:hp}
  \end{center}
\end{table}

 \begin{table}[htbp]
    \caption{Downstream task scenario configurations}
    \begin{center}
    \begin{tabular}{|l|c|c|}
    \hline
    \textbf{Parameter}&\textbf{Task seen}&\textbf{Task unseen} \\
    \hline
    Samples & $\sim$ 32K & $\sim$ 40K\\
    \hline
    Carrier frequency          & 3.5 GHz       & 3.5 GHz \\
    \hline
    {Subcarriers (OFDM)}$^{\mathrm{a}}$       & 512           & 512, 1024 \\
    \hline
    {Selected subcarriers}$^{\mathrm{a}}$       & 48, 144  & 480, 512, 256 \\
    \hline
    {BS antenna shape}$^{\mathrm{a}}$ & [4,2] and [4,4] & [8,6], [4,8] and [4,6] \\
    \hline
    {UE antenna shape}$^{\mathrm{a}}$ & [2,1]  & [2,1] and [2,2] \\
    \hline
    {BS antenna spacing}$^{\mathrm{a}}$ & 0.25$\lambda$ & 0.25$\lambda$  or 0.5$\lambda$\\
    \hline
    {UE antenna spacing}$^{\mathrm{a}}$ & 0.25$\lambda$ & 0.25$\lambda$ or 0.5$\lambda$ \\\hline
    Unique \# of $\text{CFG}$ &3&3\\
    \hline
    \multicolumn{3}{l}{$^{\mathrm{a}}$Values across the three base stations in the scenario.} \\
    \end{tabular}
    \label{tab:downstream_configs}
    \end{center}
\end{table}

\subsection{Downstream configurations}
\subsubsection{Tasks}
Three downstream tasks are evaluated on a new scenario - {\small{\texttt{city\_19\_oklahoma\_3p5\_lwm}}}: 
\begin{itemize}
\item (Non-) Line of Sight classification (LoS/NLoS) classification
\item Timing Advance (TA) estimation
\item Beam selection with codebook sizes 8, {\color{black}16, 32} (referred to as Beam-8, Beam-16 and Beam-32, respectively)
\end{itemize}
Labels are sourced from DeepMIMO, except for TA estimation where TA are computed from path delays.
\subsubsection{Scenarios}
Each task is evaluated on $H$ from this new scenario with two $\text{CFG}$ conditions: 
\begin{itemize}
    \item Seen — a $\text{CFG}$ combination sampled from pretraining configurations (Table \ref{tab:pretraining_configs})
    \item Unseen — New $\text{CFG}$ combinations not observed during pretraining (Table \ref{tab:downstream_configs}), probing for OOD generalization
\end{itemize}
Tasks are evaluated across a range of embeddings to find patterns across training data regimes. The following are considered for comparison of performance:
\begin{itemize}
\item Baseline 1, B1  - $H$ tokens, no RFM: $T_{H}$
    \item Baseline 2, B2 - $H$ tokens and $\text{CFG}$ tokens: $T_{H, \text{CFG}}$
    \item Baseline 3, B - $f^b$ RFM, no $\text{CFG}$ :  $E^b_{\text{CLS}}$, $E^b_{H_\text{mean}}$ and $E^b_{H_\text{max}}$ 
    \item Proposed, P - $f^p$ RFM, $\text{CFG}$-enriched: $E^p_{\text{CLS}}$, $E^p_{\text{CFG}_\text{mean}}$, $E^p_{\text{CFG}_{\text{max}}}$
    \item Proposed, P  - $f^p$ RFM, $\text{CFG}$-augmented: $E^p_{H, \text{CFG}_\text{mean}}$, $E^p_{H, \text{CFG}_{\text{max}}}$, $E^p_{H, \text{CLS},\text{CFG}_\text{mean}}$, $E^p_{H,\text{CLS},\text{CFG}_\text{max}}$

\end{itemize}

\subsubsection{Downstream Model Architecture}
\label{subsec:agg-windows}

A fixed FCN head (50 epochs, learning rate $10^{-3}$, batch size 8, dropout 0.3, early stopping patience 5, class-weighted loss) is used uniformly across all embeddings and for baseline 1 and baseline 3 models where raw $H$ tokens form the input. The downstream models are trained on a subset of seen $\text{CFG}$ conditions of $H$, then tested on both seen and unseen conditions of $H$.

\subsubsection{Downstream Training-Data Regimes}
Each embedding is evaluated across three training-data regimes, with performance reported as mean $\pm$ std over multiple trials:
\begin{itemize}
\item \textbf{Few-shot} $\in [10, 100]$ samples ($\sim$10 trials)
\item \textbf{Moderate} $\in (100, 1000)$ samples ($\sim$5 trials)
\item \textbf{Data-rich} $\geq 1000$ samples ($\sim$5 trials)
\end{itemize}
Unlike NLP, where millions of labelled examples are typical, wireless channel data occupies a low-dimensional structured manifold; 1000 labelled samples thus constitutes a data-rich regime for the 64-dimensional embeddings evaluated here.

\begin{table*}[t]
\centering
\caption{{Seen} - LoS/NLoS (F1$\uparrow$), TA (NMSE$\times10^{-6}$$\downarrow$), Beam-8 (F1$\uparrow$).}
\label{tab:seen_results}
\resizebox{\textwidth}{!}{%
\begin{tabular}{llrccccccccc}
\toprule
& & & \multicolumn{3}{c}{\textbf{LoS/NLoS (F1$\uparrow$)}}
  & \multicolumn{3}{c}{\textbf{TA (NMSE$\times10^{-6}$$\downarrow$)}}
  & \multicolumn{3}{c}{\textbf{Beam-8 (F1$\uparrow$)}} \\
\cmidrule(lr){4-6}\cmidrule(lr){7-9}\cmidrule(lr){10-12}
\textbf{Grp} & \textbf{Embedding} & \textbf{Dim}
  & \textbf{Few-shot} & \textbf{Moderate} & \textbf{Data-Rich}
  & \textbf{Few-shot} & \textbf{Moderate} & \textbf{Data-Rich}
  & \textbf{Few-shot} & \textbf{Moderate} & \textbf{Data-Rich} \\
\midrule
B1 & $T_H$ & 27{,}648
  & $0.564{\pm}0.136$ & $0.702{\pm}0.065$ & $0.785{\pm}0.056$
  & $9.981{\pm}0.882$ & $8.177{\pm}0.524$ & $6.861{\pm}0.382$
  & $0.210{\pm}0.048$ & $0.327{\pm}0.008$ & $0.426{\pm}0.036$ \\
B2 & $T_{H,\text{CFG}}$ & 27{,}659
  & $0.564{\pm}0.126$ & $0.654{\pm}0.068$ & $0.826{\pm}0.029$
  & $10.526{\pm}1.365$ & $8.156{\pm}0.569$ & $6.913{\pm}0.268$
  & $0.142{\pm}0.017$ & $0.283{\pm}0.037$ & $0.426{\pm}0.031$ \\
\midrule
B3 & $E_{\text{CLS}}$ & 64
  & $0.970{\pm}0.010$ & $0.980{\pm}0.003$ & $\mathbf{0.987{\pm}0.001}$
  & $5.833{\pm}2.840$ & $0.880{\pm}0.051$ & $0.696{\pm}0.101$
  & $0.170{\pm}0.024$ & $0.379{\pm}0.037$ & $\mathbf{0.558{\pm}0.041}$ \\
B3 & $E_{H_\text{mean}}$ & 64
  & $0.970{\pm}0.011$ & $0.979{\pm}0.002$ & $0.986{\pm}0.001$
  & $3.172{\pm}1.400$ & $0.952{\pm}0.185$ & $0.817{\pm}0.324$
  & $0.162{\pm}0.022$ & $0.330{\pm}0.037$ & $0.504{\pm}0.033$ \\
B3 & $E_{H_\text{max}}$ & 64
  & $0.967{\pm}0.023$ & $\mathbf{0.980{\pm}0.002}$ & $0.986{\pm}0.001$
  & $3.452{\pm}1.586$ & $1.086{\pm}0.057$ & $\mathbf{0.657{\pm}0.183}$
  & $0.169{\pm}0.011$ & $0.272{\pm}0.038$ & $0.489{\pm}0.046$ \\
\midrule
P & $E_{\text{CLS}}^\dagger$ & 64
  & $0.962{\pm}0.018$ & $\mathbf{0.980{\pm}0.002}$ & $\mathbf{0.987{\pm}0.001}$
  & $3.553{\pm}1.902$ & $0.911{\pm}0.080$ & $0.691{\pm}0.099$
  & $\mathbf{0.271{\pm}0.047}$ & $\mathbf{0.423{\pm}0.030}$ & $0.504{\pm}0.011$ \\
P & $E_{\text{CFG}_\text{mean}}$ & 64
  & $0.958{\pm}0.017$ & $0.976{\pm}0.004$ & $0.986{\pm}0.002$
  & $4.285{\pm}1.007$ & $1.162{\pm}0.284$ & $0.829{\pm}0.139$
  & $0.248{\pm}0.051$ & $0.366{\pm}0.006$ & $0.509{\pm}0.024$ \\
P & $E_{\text{CFG}_\text{max}}$ & 64
  & $0.961{\pm}0.017$ & $0.979{\pm}0.004$ & $\mathbf{0.987{\pm}0.001}$
  & $5.279{\pm}1.320$ & $1.282{\pm}0.362$ & $0.778{\pm}0.113$
  & $0.226{\pm}0.060$ & $0.397{\pm}0.008$ & $0.529{\pm}0.012$ \\
P & $E_{H,\text{CFG}_\text{mean}}$ & 64
  & $0.964{\pm}0.017$ & $0.976{\pm}0.003$ & $0.985{\pm}0.001$
  & $1.995{\pm}0.920$ & $\mathbf{0.835{\pm}0.118}$ & $0.664{\pm}0.097$
  & $0.199{\pm}0.036$ & $0.361{\pm}0.015$ & $0.501{\pm}0.014$ \\
P & $E_{H,\text{CFG}_\text{max}}$ & 64
  & $\mathbf{0.971{\pm}0.009}$ & $0.978{\pm}0.004$ & $\mathbf{0.987{\pm}0.001}$
  & $2.395{\pm}0.930$ & $1.017{\pm}0.187$ & $0.762{\pm}0.145$
  & $0.158{\pm}0.040$ & $0.327{\pm}0.035$ & $0.503{\pm}0.027$ \\
P & $E_{H,\text{CLS},\text{CFG}_\text{mean}}$ & 64
  & $0.964{\pm}0.017$ & $0.975{\pm}0.003$ & $0.986{\pm}0.001$
  & $\mathbf{1.994{\pm}0.823}$ & $0.843{\pm}0.108$ & $0.696{\pm}0.120$
  & $0.196{\pm}0.033$ & $0.359{\pm}0.017$ & $0.501{\pm}0.015$ \\
P & $E_{H,\text{CLS},\text{CFG}_\text{max}}$ & 64
  & $\mathbf{0.971{\pm}0.009}$ & $0.978{\pm}0.004$ & $\mathbf{0.987{\pm}0.001}$
  & $2.333{\pm}0.766$ & $0.984{\pm}0.189$ & $0.800{\pm}0.164$
  & $0.159{\pm}0.040$ & $0.318{\pm}0.041$ & $0.507{\pm}0.026$ \\
\bottomrule
\end{tabular}}
\end{table*}

\begin{table*}[t]
\centering
\caption{{Unseen } - LoS/NLoS (F1$\uparrow$), TA (NMSE$\times10^{-6}$$\downarrow$), Beam-8 (F1$\uparrow$).}
\label{tab:unseen_results}
\resizebox{\textwidth}{!}{%
\begin{tabular}{llrccccccccc}
\toprule
& & & \multicolumn{3}{c}{\textbf{LoS/NLoS (F1$\uparrow$)}}
  & \multicolumn{3}{c}{\textbf{TA (NMSE$\times10^{-6}$$\downarrow$)}}
  & \multicolumn{3}{c}{\textbf{Beam-8 (F1$\uparrow$)}} \\
\cmidrule(lr){4-6}\cmidrule(lr){7-9}\cmidrule(lr){10-12}
\textbf{Grp} & \textbf{Embedding} & \textbf{Dim}
  & \textbf{Few-shot} & \textbf{Moderate} & \textbf{Data-Rich}
  & \textbf{Few-shot} & \textbf{Moderate} & \textbf{Data-Rich}
  & \textbf{Few-shot} & \textbf{Moderate} & \textbf{Data-Rich} \\
\midrule
B1 & $T_H$ & 27{,}648
  & $0.348{\pm}0.046$ & $0.405{\pm}0.056$ & $0.453{\pm}0.044$
  & $15.536{\pm}3.026$ & $17.923{\pm}2.111$ & $15.714{\pm}2.349$
  & $0.166{\pm}0.064$ & $\mathbf{0.244{\pm}0.009}$ & $0.076{\pm}0.013$ \\
B2 & $T_{H,\text{CFG}}$ & 27{,}659
  & $0.348{\pm}0.039$ & $0.400{\pm}0.026$ & $0.458{\pm}0.046$
  & $15.803{\pm}3.251$ & $17.312{\pm}1.267$ & $14.540{\pm}1.272$
  & $0.118{\pm}0.020$ & $0.147{\pm}0.015$ & $0.105{\pm}0.028$ \\
\midrule
B3 & $E_{\text{CLS}}$ & 64
  & $0.887{\pm}0.051$ & $0.849{\pm}0.031$ & $0.882{\pm}0.033$
  & $2.496{\pm}2.619$ & $1.811{\pm}0.095$ & $2.620{\pm}0.279$
  & $0.182{\pm}0.013$ & $0.201{\pm}0.004$ & $0.215{\pm}0.004$ \\
B3 & $E_{H,\text{mean}}$ & 64
  & $0.918{\pm}0.068$ & $0.860{\pm}0.057$ & $0.756{\pm}0.110$
  & $\mathbf{2.456{\pm}1.754}$ & $2.428{\pm}0.371$ & $2.230{\pm}0.176$
  & $0.190{\pm}0.019$ & $0.199{\pm}0.020$ & $0.202{\pm}0.009$ \\
B3 & $E_{H,\text{max}}$ & 64
  & $0.911{\pm}0.068$ & $0.836{\pm}0.072$ & $0.762{\pm}0.103$
  & $3.161{\pm}3.311$ & $4.093{\pm}1.376$ & $2.052{\pm}0.222$
  & $0.185{\pm}0.023$ & $0.175{\pm}0.026$ & $0.220{\pm}0.008$ \\
\midrule
P & $E_{\text{CLS}}^\dagger$ & 64
  & $\mathbf{0.930{\pm}0.029}$ & $\mathbf{0.961{\pm}0.010}$ & $\mathbf{0.958{\pm}0.023}$
  & $4.458{\pm}1.812$ & $4.969{\pm}0.982$ & $9.587{\pm}1.484$
  & $\mathbf{0.211{\pm}0.019}$ & $0.175{\pm}0.011$ & $0.192{\pm}0.007$ \\
P & $E_{\text{CFG}_\text{mean}}$ & 64
  & $0.845{\pm}0.074$ & $0.864{\pm}0.071$ & $0.868{\pm}0.042$
  & $3.853{\pm}2.489$ & $\mathbf{1.735{\pm}0.180}$ & $2.154{\pm}0.190$
  & $0.154{\pm}0.039$ & $0.231{\pm}0.017$ & $0.220{\pm}0.010$ \\
P & $E_{\text{CFG}_\text{max}}$ & 64
  & $0.868{\pm}0.078$ & $0.804{\pm}0.065$ & $0.857{\pm}0.055$
  & $3.967{\pm}3.751$ & $2.004{\pm}0.206$ & $2.012{\pm}0.162$
  & $0.149{\pm}0.026$ & $0.223{\pm}0.009$ & $\mathbf{0.223{\pm}0.022}$ \\
P & $E_{H,\text{CFG}_\text{mean}}$ & 64
  & $0.855{\pm}0.081$ & $0.762{\pm}0.077$ & $0.640{\pm}0.104$
  & $2.508{\pm}1.524$ & $1.998{\pm}0.345$ & $2.420{\pm}0.258$
  & $0.199{\pm}0.013$ & $0.183{\pm}0.006$ & $0.198{\pm}0.015$ \\
P & $E_{H,\text{CFG}_\text{max}}$ & 64
  & $0.872{\pm}0.084$ & $0.935{\pm}0.032$ & $0.865{\pm}0.084$
  & $3.422{\pm}1.116$ & $2.096{\pm}0.190$ & $\mathbf{1.678{\pm}0.179}$
  & $0.159{\pm}0.019$ & $0.199{\pm}0.007$ & $0.183{\pm}0.014$ \\
P & $E_{H,\text{CLS}\text{CFG}_\text{mean}}$ & 64
  & $0.856{\pm}0.084$ & $0.742{\pm}0.084$ & $0.634{\pm}0.096$
  & $2.499{\pm}1.404$ & $2.032{\pm}0.218$ & $2.246{\pm}0.127$
  & $0.199{\pm}0.013$ & $0.183{\pm}0.007$ & $0.199{\pm}0.015$ \\
P & $E_{H,\text{CLS},\text{CFG}_\text{max}}$ & 64
  & $0.875{\pm}0.079$ & $0.930{\pm}0.027$ & $0.848{\pm}0.100$
  & $3.321{\pm}1.209$ & $2.156{\pm}0.388$ & $\mathbf{1.674{\pm}0.110}$
  & $0.161{\pm}0.017$ & $0.199{\pm}0.006$ & $0.183{\pm}0.014$ \\
\bottomrule
\end{tabular}}
\end{table*}

\begin{table*}[t]
\centering
\caption{{Seen} - Beam-16 and Beam-32 (F1$\uparrow$).}
\label{tab:seen_beam_16_32}
\resizebox{0.75\textwidth}{!}{%
\begin{tabular}{llrcccccc}
\toprule
& & & \multicolumn{3}{c}{\textbf{Beam-16 (F1$\uparrow$)}}
  & \multicolumn{3}{c}{\textbf{Beam-32 (F1$\uparrow$)}} \\
\cmidrule(lr){4-6}\cmidrule(lr){7-9}
\textbf{Grp} & \textbf{Embedding} & \textbf{Dim}
  & \textbf{Low} & \textbf{Med} & \textbf{Rich}
  & \textbf{Low} & \textbf{Med} & \textbf{Rich} \\
\midrule
B1 & $T_H$ & 27{,}648
  & $0.076{\pm}0.017$ & $0.095{\pm}0.024$ & $0.256{\pm}0.091$
  & $0.019{\pm}0.006$ & $0.052{\pm}0.022$ & $0.212{\pm}0.053$ \\
B2 & $T_{H,\text{CFG}}$ & 27{,}659
  & $0.039{\pm}0.015$ & $0.068{\pm}0.030$ & $0.331{\pm}0.049$
  & $0.027{\pm}0.007$ & $0.038{\pm}0.009$ & $0.187{\pm}0.075$ \\
\midrule
B3 & $E_{\text{CLS}}$ & 64
  & $0.067{\pm}0.029$ & $0.176{\pm}0.027$ & $0.446{\pm}0.047$
  & $0.018{\pm}0.014$ & $0.182{\pm}0.041$ & $\mathbf{0.356{\pm}0.045}$ \\
B3 & $E_{H_\text{mean}}$ & 64
  & $0.070{\pm}0.041$ & $0.166{\pm}0.023$ & $0.390{\pm}0.050$
  & $0.042{\pm}0.044$ & $0.163{\pm}0.011$ & $0.266{\pm}0.050$ \\
B3 & $E_{H_\text{max}}$ & 64
  & $0.072{\pm}0.048$ & $0.158{\pm}0.022$ & $0.360{\pm}0.063$
  & $0.065{\pm}0.027$ & $0.178{\pm}0.015$ & $0.237{\pm}0.025$ \\
\midrule
P & $E_{\text{CLS}}^\dagger$ & 64
  & $0.052{\pm}0.026$ & $\mathbf{0.254{\pm}0.030}$ & $0.409{\pm}0.030$
  & $0.049{\pm}0.058$ & $0.220{\pm}0.028$ & $0.352{\pm}0.028$ \\
P & $E_{\text{CFG}_\text{mean}}$ & 64
  & $0.074{\pm}0.016$ & $0.213{\pm}0.027$ & $0.379{\pm}0.031$
  & $0.034{\pm}0.032$ & $0.194{\pm}0.048$ & $0.304{\pm}0.052$ \\
P & $E_{\text{CFG}_\text{max}}$ & 64
  & $\mathbf{0.097{\pm}0.020}$ & $0.239{\pm}0.030$ & $\mathbf{0.454{\pm}0.033}$
  & $\mathbf{0.093{\pm}0.055}$ & $\mathbf{0.236{\pm}0.028}$ & $0.350{\pm}0.051$ \\
P & $E_{H,\text{CFG}_\text{mean}}$ & 64
  & $0.056{\pm}0.032$ & $0.158{\pm}0.036$ & $0.366{\pm}0.057$
  & $0.034{\pm}0.045$ & $0.198{\pm}0.027$ & $0.295{\pm}0.033$ \\
P & $E_{H,\text{CFG}_\text{max}}$ & 64
  & $0.056{\pm}0.022$ & $0.151{\pm}0.030$ & $0.356{\pm}0.066$
  & $0.048{\pm}0.045$ & $0.192{\pm}0.017$ & $0.280{\pm}0.027$ \\
P & $E_{H,\text{CLS},\text{CFG}_\text{mean}}$ & 64
  & $0.056{\pm}0.032$ & $0.158{\pm}0.036$ & $0.364{\pm}0.052$
  & $0.034{\pm}0.044$ & $0.199{\pm}0.026$ & $0.299{\pm}0.041$ \\
P & $E_{H,\text{CLS},\text{CFG}_\text{max}}$ & 64
  & $0.056{\pm}0.022$ & $0.148{\pm}0.029$ & $0.374{\pm}0.055$
  & $0.047{\pm}0.046$ & $0.191{\pm}0.016$ & $0.285{\pm}0.031$ \\
\bottomrule
\end{tabular}}
\end{table*}

\begin{table*}[!t]
\centering
\caption{{Unseen} - Beam-16 and Beam-32 (F1$\uparrow$).}
\label{tab:unseen_beam_16_32}
\resizebox{0.75\textwidth}{!}{%
\begin{tabular}{llrcccccc}
\toprule
& & & \multicolumn{3}{c}{\textbf{Beam-16 (F1$\uparrow$)}}
  & \multicolumn{3}{c}{\textbf{Beam-32 (F1$\uparrow$)}} \\
\cmidrule(lr){4-6}\cmidrule(lr){7-9}
\textbf{Grp} & \textbf{Embedding} & \textbf{Dim}
  & \textbf{Low} & \textbf{Med} & \textbf{Rich}
  & \textbf{Low} & \textbf{Med} & \textbf{Rich} \\
\midrule
B1 & $T_H$ & 27{,}648
  & $0.049{\pm}0.013$ & $0.077{\pm}0.056$ & $0.117{\pm}0.054$
  & $0.010{\pm}0.005$ & $0.019{\pm}0.003$ & $0.018{\pm}0.003$ \\
B2 & $T_{H,\text{CFG}}$ & 27{,}659
  & $0.032{\pm}0.009$ & $0.034{\pm}0.011$ & $0.102{\pm}0.040$
  & $0.012{\pm}0.004$ & $0.041{\pm}0.041$ & $0.028{\pm}0.013$ \\
\midrule
B3 & $E_{\text{CLS}}$ & 64
  & $0.038{\pm}0.014$ & $0.140{\pm}0.027$ & $\mathbf{0.130{\pm}0.013}$
  & $0.015{\pm}0.001$ & $0.016{\pm}0.001$ & $0.026{\pm}0.005$ \\
B3 & $E_{H_\text{mean}}$ & 64
  & $0.049{\pm}0.036$ & $0.130{\pm}0.039$ & $0.110{\pm}0.012$
  & $0.014{\pm}0.001$ & $0.017{\pm}0.011$ & $0.010{\pm}0.004$ \\
B3 & $E_{H_\text{max}}$ & 64
  & $\mathbf{0.103{\pm}0.021}$ & $\mathbf{0.155{\pm}0.023}$ & $0.129{\pm}0.011$
  & $\mathbf{0.036{\pm}0.011}$ & $\mathbf{0.053{\pm}0.003}$ & $\mathbf{0.057{\pm}0.011}$ \\
\midrule
P & $E_{\text{CLS}}^\dagger$ & 64
  & $0.043{\pm}0.048$ & $0.047{\pm}0.012$ & $0.052{\pm}0.014$
  & $0.011{\pm}0.003$ & $0.010{\pm}0.002$ & $0.008{\pm}0.001$ \\
P & $E_{\text{CFG}_\text{mean}}$ & 64
  & $0.054{\pm}0.035$ & $0.095{\pm}0.011$ & $0.089{\pm}0.017$
  & $0.007{\pm}0.003$ & $0.006{\pm}0.002$ & $0.004{\pm}0.001$ \\
P & $E_{\text{CFG}_\text{max}}$ & 64
  & $0.091{\pm}0.050$ & $0.115{\pm}0.022$ & $0.109{\pm}0.008$
  & $0.011{\pm}0.002$ & $0.005{\pm}0.002$ & $0.004{\pm}0.001$ \\
P & $E_{H,\text{CFG}_\text{mean}}$ & 64
  & $0.034{\pm}0.049$ & $0.078{\pm}0.032$ & $0.014{\pm}0.010$
  & $0.004{\pm}0.002$ & $0.002{\pm}0.001$ & $0.003{\pm}0.001$ \\
P & $E_{H,\text{CFG}_\text{max}}$ & 64
  & $0.046{\pm}0.037$ & $0.124{\pm}0.036$ & $0.081{\pm}0.008$
  & $0.006{\pm}0.003$ & $0.009{\pm}0.002$ & $0.012{\pm}0.004$ \\
P & $E_{H,\text{CLS},\text{CFG}_\text{mean}}$ & 64
  & $0.032{\pm}0.045$ & $0.079{\pm}0.029$ & $0.014{\pm}0.010$
  & $0.004{\pm}0.002$ & $0.002{\pm}0.000$ & $0.003{\pm}0.001$ \\
P & $E_{H,\text{CLS},\text{CFG}_\text{max}}$ & 64
  & $0.050{\pm}0.036$ & $0.122{\pm}0.035$ & $0.080{\pm}0.010$
  & $0.006{\pm}0.003$ & $0.009{\pm}0.002$ & $0.012{\pm}0.003$ \\
\bottomrule
\end{tabular}}
\end{table*}
\subsection{Results and Analysis}
Table~\ref{tab:seen_results} and \ref{tab:seen_beam_16_32} consolidates the performance of all embeddings across three data regimes for all tasks: LoS/NLoS classification, TA estimation, Beam-8, Beam-16 and Beam-32 selection on seen $\text{CFG}$ scenarios. 

Results are presented as mean $\pm$ std over all training sizes within each regime, with bold indicating the best-performing embedding per task and regime. 
For the unseen $\text{CFG}$ scenarios, the performance is aggregated in Table~\ref{tab:unseen_results} and \ref{tab:unseen_beam_16_32}. General patterns observed are as follows: 
\begin{itemize}
    \item Seen scenario - performance generally improves with data size - which is expected.
    \item Unseen scenario - above trend does not generally hold; performance reflects the downstream model's ability to generalise (to the test set), as a function of training data size.
\end{itemize}

\subsubsection{LoS/NLoS classification}
Weighted-F1 are reported for this binary classification task (refer Table~\ref{tab:seen_results} and Table~\ref{tab:unseen_results}) across different training regimes the baselines (B1, B2 and B3), embeddings from the proposed models (P), with dimensionality listed. Some of the observations are as follows:
\\
\underline{B1 and B2 baselines (raw tokens):}
\begin{itemize}
\item Performance of $T_H$ and $T_{H,\text{CFG}}$ baselines is much below all 64-dimensional embeddings across all regimes. The gap widens in the unseen case. This confirms   poor OOD transferability for LoS/NLoS classification.
\item Appending $T_\text{CFG}$ tokens (B2 vs.\ B1) yields a modest gain in the seen data-rich regime but negligible improvement in the unseen regime, indicating that $T_\text{CFG}$ do not improve OOD generalisation.
\end{itemize}
\underline{RFM embeddings (B3 and P):}
\begin{itemize}
    \item Performance saturates rapidly across training sizes in the seen scenario. It indicates low training sizes suffice for near-peak performance. B3 and P embeddings are highly competitive, with differences $\leq 0.001$ F1 for all regimes.
    \item In the seen scenario, $\text{CFG}$-pooled embeddings are indistinguishable from
          $E_{\text{CLS}}^\dagger$. 
    \item In the unseen scenario, $E_{\text{CLS}}^\dagger$ consistently outperforms all embedding for all data-regimes by a significant F1-score, demonstrating superior OOD robustness of the joint $H-\text{CFG}$ representation.
 
    \item Performance variance is low across all regimes and both seen and unseen configurations, reflecting stable saturation.
\end{itemize}

\subsubsection{TA Estimation}

From  (refer Table~\ref{tab:seen_results} and~\ref{tab:unseen_results}), the following patterns can be observed through analysis of aggregated TA-prediction NMSE (lower is better) across various training-size windows:

\underline{B1 and B2 baselines (raw tokens):}
\begin{itemize}
    \item Raw token baselines remain substantially inferior to all 64-dimensional embeddings across all regimes. The gap is especially severe in the unseen scenarios (B1/B2 vs B3 and P embeddings), confirming that raw tokens have negligible OOD transferability for TA regression.
    \item Adding $T_\text{CFG}$ tokens (B2 vs.\ B1) yields no consistent improvement in either seen or unseen scenarios.
    
\end{itemize}
\underline{RFM embeddings (B3 and P):}
\begin{itemize}
    \item In the seen scenario, most of P embeddings consistently outperform the best B3 embedding in the few-shot regime by a substantial margin. P is marginally better in moderate regime and remains competitive in the data-rich regime. As  training data increases, differences becoming marginal.

    \item In the unseen scenario, embeddings from B3 and P are comparable in the few-shot and moderate regimes. In the data-rich
          regime, proposed channel-augmented embeddings match or outperform B3.

    \item Channel-augmented aggregate embeddings, $\text{Agg*}$\footnote{$\text{Agg*}$ to indicate mean/max pooling} from P (i.e., $E_{H, \text{CFG}~\text{Agg*}}$, $E_{H,\text{CLS},\text{CFG}~\text{Agg*}}$)
          consistently outperform CFG-only pooled embeddings ($E_{\text{CFG}~\text{Agg*}}$) across all regimes in the seen scenario; this confirms that $H$ information is
          essential for TA regression, as TA is a function of the ToA extractable from the channel impulse response. $E_\text{CFG}$ adds value but cannot substitute for $E_H$.
    \item The few-shot regime exhibits high variance for nearly all embeddings on both seen and
          unseen configurations, reflecting the inherent instability of TA regression with very
          few labelled samples.
        \end{itemize}

\subsubsection{Beam Selection: Beam-8}
Notable observations (refer Table~\ref{tab:seen_results} and Table~\ref{tab:unseen_results}) from the mean $\pm$ std of weighted F1-scores  for Beam-8 task are: 

\underline{B1 and B2 baselines (raw tokens):}
\begin{itemize}
    \item $T_\text{CFG}$ augmentation (B2 vs.\ B1) provides no benefit across any regime or
          scenario, with B2 consistently equal to or worse than B1.
    \item In the unseen scenario, both B1 and B2 collapse and remain flat across all training
          sizes, demonstrating severe OOD generalisation failure.
\end{itemize}

\underline{RFM embeddings (B3 and P):}
\begin{itemize}
    \item In the seen scenario, P embeddings consistently outperform B3 in the few-shot and moderate
          regimes, with $E_{\text{CLS}}^\dagger$ leading; also, larger gains at smaller training
          sizes are observed. In the data-rich regime, B3 $E_\text{CLS}$ outperforms other RFM embeddings.
    \item The dominance of $E_{\text{CLS}}^\dagger$ under label scarcity suggests pooling incurs
          a performance cost for this task; spatially richer embeddings are better suited when training data is limited;  data abundance in the rich regime compensates for
          pooling losses. 
    \item In the unseen scenario, P embeddings ($E_{\text{CLS}}^\dagger$, $E_{\text{CFG,max}}$)
          lead in the few-shot and data-rich regimes; in the moderate regime B1 occasionally edges ahead,
          though the best P embedding remains competitive.
    \item The seen-to-unseen gap grows substantially with training size, consistent with greater
          overfitting at larger sizes --- making few-shot regime performance relatively more
          stable under distribution shift.
    \item All RFM embeddings exhibit low variance across training sizes in the unseen scenario,
          indicating consistent beam selection performance even under OOD conditions.
\end{itemize}


\subsubsection{Beam Selection: Beam-16 and Beam-32}
Beam-16 and Beam-32 outcomes for seen and unseen $\text{CFG}$ are summarized in \ref{tab:seen_beam_16_32} and \ref{tab:unseen_beam_16_32}; trends are broadly consistent with Beam-8 in the seen scenario, with proposed embeddings ($E_{\text{CFG},\text{max}}$, $E_{\text{CLS}}^\dagger$) leading in few-shot and moderate regimes and B3 $E_{\text{CLS}}$ in the data-rich regime. In the unseen scenario, however, $E_{H_\text{max}}$ (B3) dominates across all regimes for both tasks, with proposed embeddings failing to match B3 --- suggesting $\text{CFG}$ enrichment does not confer OOD robustness for finer-grained beam discrimination. Overall F1 scores are substantially lower for Beam-16 and Beam-32 (than Beam-8), reflecting the increased difficulty with more complex beam sets.

\subsection{Summary}
Overall the findings can be summarized as:
\begin{itemize}
    \item Across all the downstream tasks, three training regimes, and both seen and OOD
          scenarios, embeddings containing $\text{CFG}$ achieve the best or competitive performance against
          all baselines, with gains more pronounced in OOD settings.
    \item Task type modulates the utility of CFG: LoS/NLoS and beam selection benefit more
          consistently than TA prediction, where $H$ remains the dominant factor.
    \item For beam selection, performance degrades with increasing beam count (8$\to$16$\to$32); advantages observed in seen scenarios by introducing $\text{CFG}$
          will diminish in OOD settings for
          finer-grained, complex beam discrimination task.
    \item CFG-enriched embeddings ($E_{\text{CLS}}^\dagger$, $E_{\text{CFG,Agg}}$) tend to
          outperform CFG-augmented embeddings ($E_{H,\text{CFG,Agg}}$, $E_{\text{CLS},H,\text{CFG,Agg}}$)
          in few-shot and moderate data regimes;  augmented embeddings are more competitive in the
          data-rich regime --- suggesting that effective $\text{CFG}$-augmentation requires greater training
          data, whereas $\text{CFG}$-enriched embeddings provide more value under label scarcity.
\end{itemize}

\subsection{T-SNE visualization of embeddings}
\begin{figure*}[t]
        \includegraphics[width=0.48\linewidth]{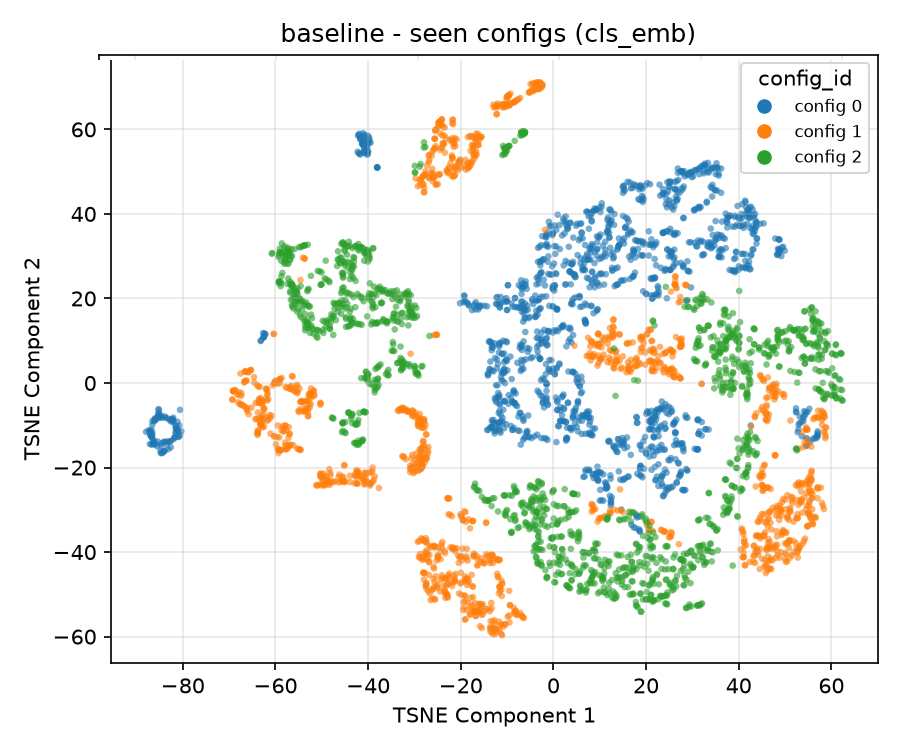}
        \includegraphics[width=0.48\linewidth]{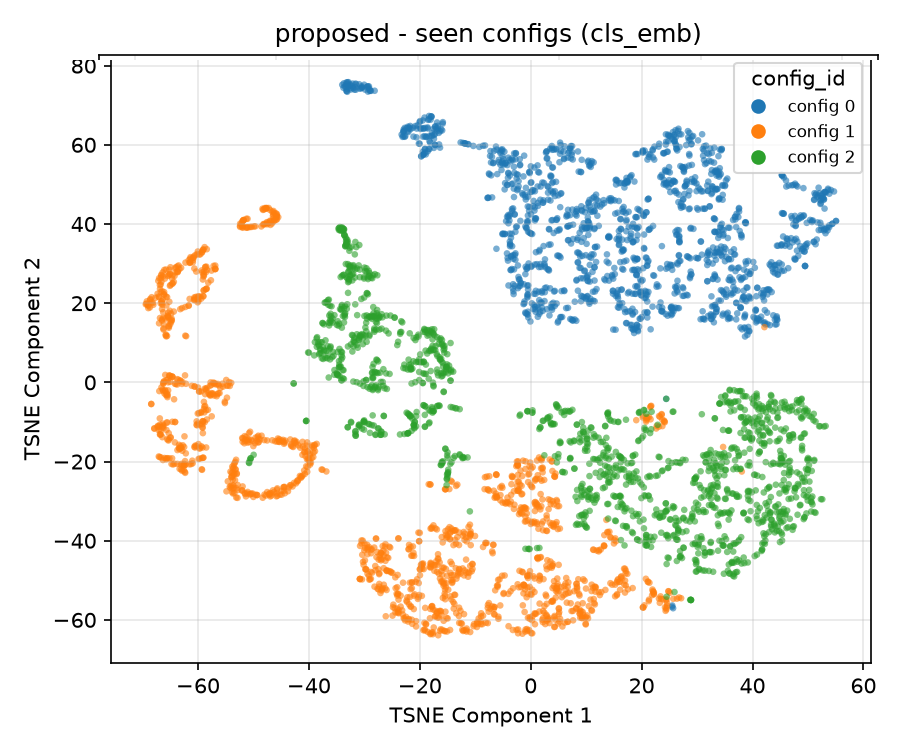}
        \caption{A t-SNE visualization of 64-dimensional (a) $E_{\text{CLS}}$ embeddings from baseline 3 (B3) model (b) $E_{\text{CLS}}^\dagger$ embeddings from proposed approach from the seen configurations. }
        \label{fig:tsne_seen}
        \includegraphics[width=0.48\linewidth]{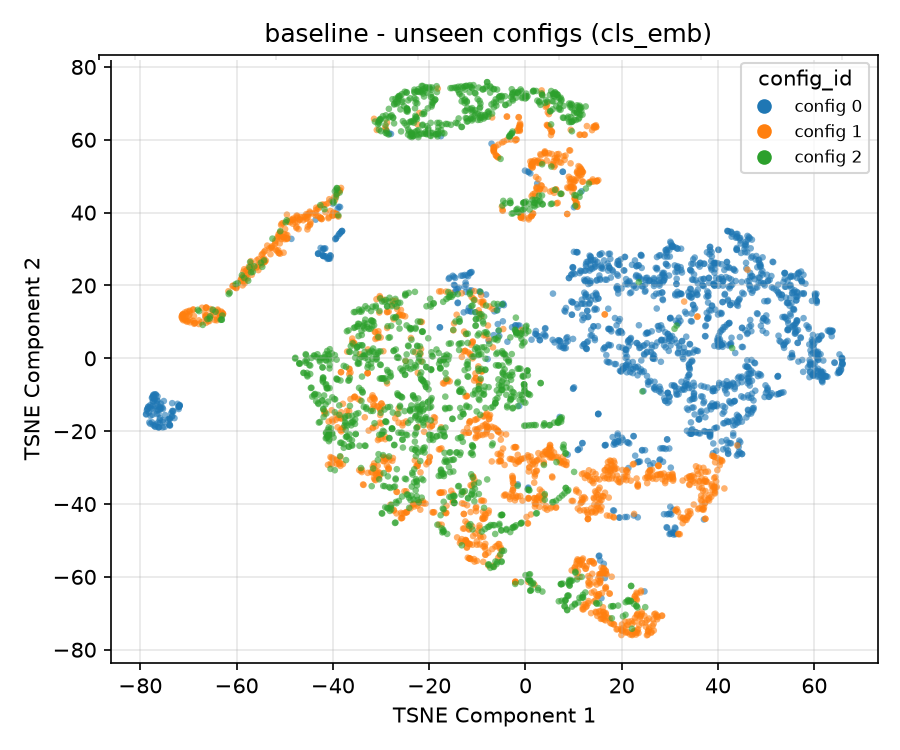}
        \includegraphics[width=0.48\linewidth]{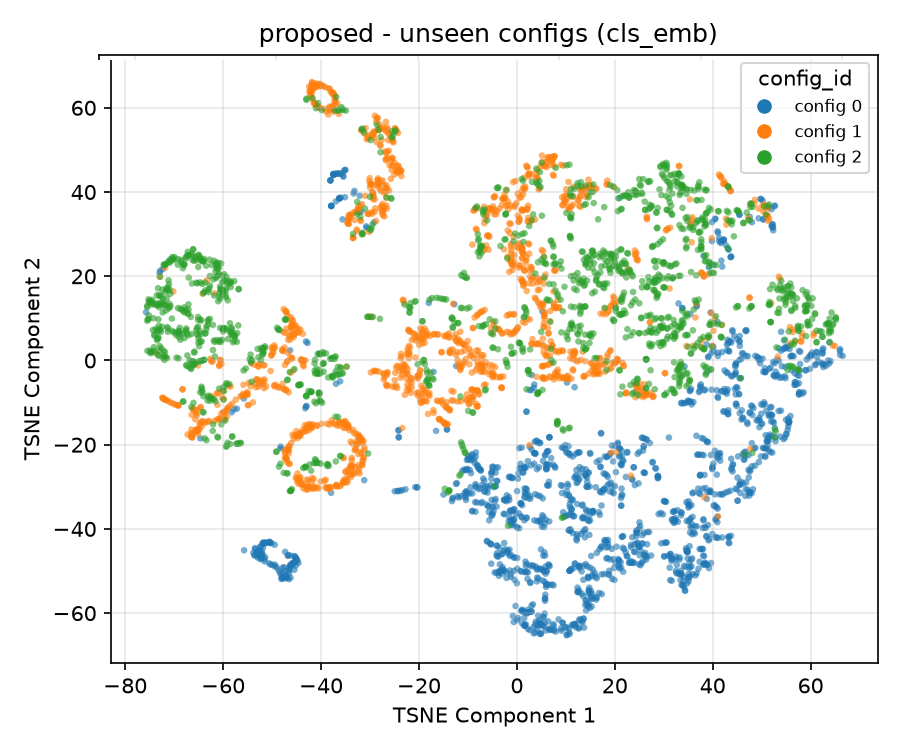}
        \caption{A t-SNE visualization of 64-dimensional (a) $E_{\text{CLS}}$ embeddings from baseline 3 (B3) model (b) $E_{\text{CLS}}^\dagger$ embeddings from proposed approach from the unseen configurations. }
        \label{fig:tsne_unseen}
\end{figure*}

$\text{CFG}$-enriched embeddings exhibit clearer cluster separation in t-SNE visualisations (refer Figure \ref{fig:tsne_seen} and \ref{fig:tsne_unseen}), indicating that conditioning on propagation geometry encourages the encoder to organise the latent space along semantically meaningful axes. This improved structure suggests that $\text{CFG}$ tokens act as an inductive prior that pulls same-class representations closer while pushing distinct propagation regimes apart, directly translating to more generalisable downstream decision boundaries.

\section{Conclusion}
Experimental outcomes confirm that compact 64-dimensional $\text{CFG}$-enriched embeddings consistently outperform high-dimensional raw channel tokens across all three tasks and training regimes, particularly in few-shot and OOD regimes. $E_{\text{CLS}}^\dagger$ demonstrates robust OOD generalisation for classification. Gains are less consistent for TA estimation under moderate OOD, and B3 $E_{\text{CLS}}$ remains competitive in data-rich beam selection. The consistent benefit of $\text{CFG}$ tokens suggests they encode propagation-relevant structure acting as an inductive prior, improving downstream generalisation. $\text{CFG}$ tokens encode coarse propagation geometry sufficient for binary or moderate-resolution tasks, but lack the fine spatial resolution needed when adjacent beams differ by only a few degrees (Beam-16 and Beam-32). As part of future work, it would be of interest to extend the $\text{CFG}$ tokens to richer descriptors to broaden the inductive prior beyond geometry. On the modelling front, one can use contrastive learning to define positive pairs (channels sharing geometry or environment type) thus enabling invariant propagation structure to emerge without task labels.


\end{document}